\documentclass[runningheads]{llncs}
\usepackage{graphicx}
\usepackage{amsmath,amssymb,bm}
\usepackage{mathtools}

\usepackage[linesnumbered, ruled, vlined]{algorithm2e}

\usepackage{xcolor}
\usepackage{tikz}
\usetikzlibrary{calc, shapes, fit}
\usepackage{pgfplots}
\pgfplotsset{compat=1.10}

\definecolor{tucol1}{rgb}{1.0 0.5 0.05}
\definecolor{tucol2}{rgb}{0.52,0.72,0.10}
\definecolor{tucol3}{rgb}{0.0 0.0 0.8}
\definecolor{tucol5}{rgb}{0.8 0.8 0}
\definecolor{tucol4}{rgb}{0.8 0 0.8}
\definecolor{tucol6}{rgb}{0 0.8 0.8}

\definecolor{tucol-kraeftig1}{RGB}{132,184,25}
\definecolor{tucol-kraeftig2}{RGB}{216,148,39}
\definecolor{tucol-kraeftig3}{RGB}{27,161,175}
\definecolor{tucol-kraeftig4}{RGB}{168,0,135}
\definecolor{tucol-kraeftig5}{RGB}{239,228,32}
\definecolor{tucol-kraeftig6}{RGB}{202,116,40}

\definecolor{tucol-kuehl1}{RGB}{132,184,25}
\definecolor{tucol-kuehl2}{RGB}{233,238,168}
\definecolor{tucol-kuehl3}{RGB}{217,233,229}
\definecolor{tucol-kuehl4}{RGB}{191,218,187}
\definecolor{tucol-kuehl5}{RGB}{113,176,96}
\definecolor{tucol-kuehl6}{RGB}{211,223,99}

\definecolor{tucol-warm1}{RGB}{132,184,25}
\definecolor{tucol-warm2}{RGB}{112,61,46}
\definecolor{tucol-warm3}{RGB}{193,178,40}
\definecolor{tucol-warm4}{RGB}{228,200,38}
\definecolor{tucol-warm5}{RGB}{178,175,132}
\definecolor{tucol-warm6}{RGB}{165,134,83}

\definecolor{tucol-rot1}{RGB}{132,184,25}
\definecolor{tucol-rot2}{RGB}{196,21,58}
\definecolor{tucol-rot3}{RGB}{157,19,42}
\definecolor{tucol-rot4}{RGB}{119,24,40}
\definecolor{tucol-rot5}{RGB}{97,126,31}
\definecolor{tucol-rot6}{RGB}{58,82,11}

\definecolor{tucol-gruen1}{RGB}{132,184,25}
\definecolor{tucol-gruen2}{RGB}{214,223,42}
\definecolor{tucol-gruen3}{RGB}{97,134,39}
\definecolor{tucol-gruen4}{RGB}{148,164,33}
\definecolor{tucol-gruen5}{RGB}{182,201,48}
\definecolor{tucol-gruen6}{RGB}{115,158,64}

\definecolor{tucol-blau1}{RGB}{132,184,25}
\definecolor{tucol-blau2}{RGB}{11,161,226}
\definecolor{tucol-blau3}{RGB}{36,123,196}
\definecolor{tucol-blau4}{RGB}{40,140,141}
\definecolor{tucol-blau5}{RGB}{177,213,230}
\definecolor{tucol-blau6}{RGB}{13,75,127}

\usepackage{hyphenat}
\usepackage{tabularx}
\usepackage{multirow}
\usepackage{makecell}
\usepackage{booktabs}
\usepackage{cite}
\newcolumntype{L}[1]{>{\hsize=#1\hsize\raggedright\arraybackslash}X}
\newcolumntype{R}[1]{>{\hsize=#1\hsize\raggedleft\arraybackslash}X}
\newcolumntype{C}[1]{>{\hsize=#1\hsize\centering\arraybackslash}X}

\usepackage[bookmarks=False,
colorlinks,
urlcolor=blue,
citecolor=black!50!green,
linkcolor=blue
]{hyperref}

\usepackage{cleveref}

\begin{document}
\title{CM1 - A Dataset for Evaluating Few-Shot Information Extraction with Large Vision Language Models}
\titlerunning{CM1 - A Dataset for Evaluating Few-Shot Information Extraction}
%
\author{
Fabian Wolf\inst{1,2}\orcidID{0000-0001-8842-3718}
Oliver Tüselmann\inst{1,2}\orcidID{0000-0002-8892-3306} \and \\
Arthur Matei\inst{1}\orcidID{0009-0009-6028-7502} \and 
Lukas Hennies\inst{3}\orcidID{0000-0003-1516-8183} \and
Christoph Rass\inst{3}\orcidID{0000-0001-9492-907X} \and \\
Gernot A. Fink\inst{1,2}\orcidID{0000-0002-7446-7813}}
\authorrunning{F. Wolf et al.}
%
\institute{TU Dortmund University, Department of Computer Science, \and
LAMARR, Institute for Machine Learning and Artificial Intelligence, \\ Dortmund, Germany \\ 
\email{\{firstname.lastname\}@cs.tu-dortmund.de} \and
Osnabrück University, Chair of Modern History and Historical Migration Studies, \\ Osnabrück,  Germany \\
\email{\{firstname.lastname\}@uni-osnabrueck.de}}


\maketitle              
\begin{abstract}
The automatic extraction of key-value information from handwritten documents is a key challenge in document analysis.
A reliable extraction is a prerequisite for the mass digitization efforts of many archives.
Large Vision Language Models (LVLM) are a promising technology to tackle this problem especially in scenarios where little annotated training data is available.
In this work, we present a novel dataset specifically designed to evaluate the few-shot capabilities of LVLMs.
The CM1 documents are a historic collection of forms with handwritten entries created in Europe to administer the Care and Maintenance program after World War Two.
The dataset establishes three benchmarks on extracting name and birthdate information and, furthermore, considers different training set sizes.
We provide baseline results for two different LVLMs and compare performances to an established full-page extraction model.
While the traditional full-page model achieves highly competitive performances, our experiments show that when only a few training samples are available the considered LVLMs benefit from their size and heavy pretraining and outperform the classical approach.
The dataset and instructions are available under \url{github.com/AvailableAfterReview}.
\end{abstract}

\section{Introduction}
Over most parts of the 19th and 20th century, forms and register cards built the backbone of many administrative processes.
Machine printed forms with handwritten entries were created in large quantities.
Many archives preserved huge document collections in physical form, leaving the contained information largely inaccessible.
The digitization of these collections is highly desirable.
Besides conservation, researchers from the humanities are highly interested in the information contained in the collections.
Analyzing historic and social developments at scale based on quantitative methods allows for reliable insights.
In order to statistically analyze the information in a collection with potentially millions of documents, automatic extraction methods are absolutely essential.
Already the first stage of digitizing a collection poses a tremendous problem for an archive.
Besides the physical process of scanning the documents, basic information has to be captured to index the document.
Personal information is especially valuable not just for indexing but also for historians, social scientists and genealogists.
For historic documents, the reliable identification of birthdates is quintessential as they often determine whether a document may be published.

Many approaches for information extraction rely on pipelines including multiple steps such as preprocessing, layout analysis and text recognition.
Using multiple independent models makes the approach prone to error propagations and also commonly relies on annotated training material for each of the stages \cite{Boros2020-ACO, Tarride2022-ACS}.
Recently, full-page information extraction models that allow for end-to-end learning, have become increasingly popular.
Models such as DONUT \cite{Kim2022-OFD}, Dessurt \cite{Davis2022-DESSURT}, Pix2Struct\cite{Lee2023-Pix2Struct} or DAN \cite{Coquenet2023-DAS} rely on attention mechanisms and generative text approaches such as BART \cite{Lewis2020-BART}, to extract information in a structured format.
Given enough representative and annotated training data, several works show high performances when solving the extraction of key-value pairs from document images.
Recently, Large Language Models (LLMs) have emerged and showed incredible capabilities for all kind of tasks.
Large Vision Language Models (LVLMs) such as ChatGPT4o, PaliGemma \cite{Mesnard2024-GEMMA} or Qwen \cite{Yang2024-QLL} essentially allow the generation of a textual answer, given a prompt and an image.
The task of extracting key-value information to access and index large historic collections, can be easily formulated as a prompt to a LVLM.
Recent results show that the intensive training process of LLMs makes them exceptionally well-suited for few-shot learning \cite{Brown2020_LMF}.

The amount of required training data is an essential consideration, as it has to be created manually in a labor intensive process.
On the other hand, LVLMs pose a tremendous requirement in terms of computational resources.
In this work, we present a novel dataset that simulates the situation many archives are confronted with.
The CM1 documents are machine printed forms with handwritten entries that have been created in the middle of the 20th century \cite{Borggrafe2020-EPF, Rass2020-NAF, Huhn2025-DP}.
Only few datasets exist that consider the extraction of specific key information from a full document page, especially with respect to handwritten entries.
This constitutes a very challenging task as the model has to solve the layout problem as well as it has to perform handwriting recognition.
Besides establishing several benchmarks, we investigate the training data requirements of a traditional full-page extraction model and compare performances to two open weight LVLMs.

This work is organized as follows.
\Cref{sec:cm1dataset} present the historic document collection and defines the different benchmarks that we propose to evaluate the few-shot performances of extraction models.
\Cref{sec:models} introduces the traditional full-page extraction model and two open source LVLMs, which we consider in the later experimental evaluation.
Finally, \Cref{sec:experiments} presents the performances of the different models on the newly introduced benchmarks.

\section{CM1 Dataset}
\label{sec:cm1dataset}
The documents of the CM1 dataset were created after World War Two.
When the Allies closed in on Nazi Germany in mid-1944, they anticipated encountering hundreds of thousands of deported survivors of Nazi persecution \cite{Rass2020-NAF}.
This humanitarian catastrophe prompted the emergence of a new category - "displaced persons" - to describe those uprooted by war and terror who would need assistance far from their pre-war homes \cite{Huhn2025-DP}.
Approximately 65 million people were forced to leave homes due to the war.
Of these, around 11 million were classified as displaced persons (DPs) in Europe, with 8 million located in Germany.
Most of these war victims were quickly repatriated after the end of hostilities.
In 1947, when responsibilities for the remaining DPs passed to the International Refugee Organization (IRO), about 700.000 of them were still present in Western Europe.
Today, roughly 350.000 of their case files still exist and are preserved by the Arolsen Archives \footnote[1]{\url{https://collections.arolsen-archives.org/en/archive/3-2-1}}.
 
The CM1 ("Care and Maintenance") files created by the IRO between 1947 and 1951 are among the most significant historical sources documenting this displacement crisis and its management.
These documents record complex negotiations between institutions and individuals regarding status, mobility options, and life chances in the aftermath of war.
They contain data on the applicants regarding their personal information, educational background, employment history prior to displacement, as well as information about relatives, places of residence, vocational training, and language skills.

The CM1 files allow researchers to explore pathways of forced migration, institutional decision-making patterns, and DP agency \cite{Borggrafe2020-EPF}.
Through these documents, we can observe how categories like "refugee" or "displaced person" operated in practice rather than merely as normative constructs.
This perspective aligns with the reflexive turn in migration studies, which questions the production of categories as social practices.
As archives digitize these materials, new methodological approaches become possible.
The combined use of qualitative analysis and mass data extraction enables researchers to examine both individual experiences and broader patterns, challenging conventional narratives of forced migration and illuminating the multifaceted negotiation of its aftermath.
This requires not only primary digitization resulting in digital facsimiles, but making these files machine-readable to open their data to systematic analysis.

\subsection{Data Preparation}
\begin{figure}[t]
  \includegraphics[width=\textwidth]{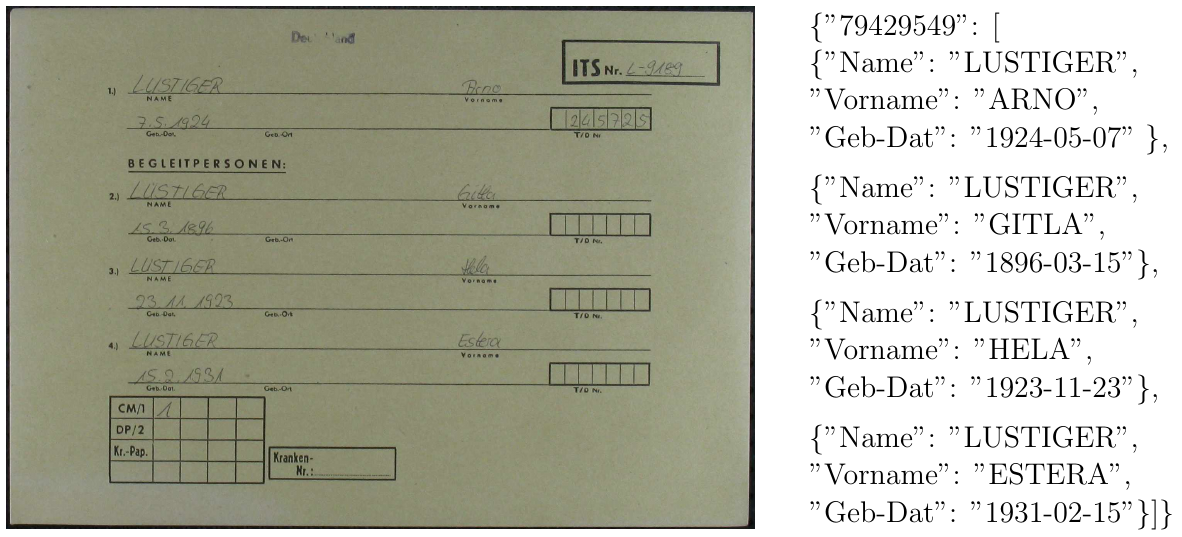}
  \vspace{-0.6cm}
  \caption{\textit{CM1-COVER} example image with corresponding JSON annotation.}
  \label{fig:mantel_beispiel}
\end{figure}

\begin{figure}[t]
  \includegraphics[width=\textwidth]{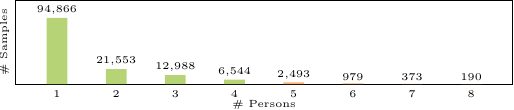}
  \vspace{-0.5cm}
  \caption{Distribution over the number of people on the cover pages. The proposed dataset only includes examples with four or less individuals.}
  \label{fig:mantel_distribution}
\end{figure}

Scans of the CM1 documents are provided by the Arolsen Archives.
The documents are organized by a distinct process identifier, which represents all documents related to the application for assistance.
One application may include multiple people filing for assistance.
Nonetheless, every process has a defined head of the family, which constitute the main person the documents are related to. 
The main purpose of the provided dataset is the evaluation of models that allow the extraction of the main and associated persons, as well as their corresponding dates of birth.
This information has been manually annotated by the archive and is publicly available.
Every process consists of several documents, including a cover page, the application form and further accompanying personal documents.
The documents are written in several different languages such as English, German, French, Polish and others.
The proposed benchmark only includes tasks related to the cover and the first page of the application.

The cover pages were not created at the same time as the application documents, as they serve organizational purposes of the archive.
Every page follows the exact same layout, see \Cref{fig:mantel_beispiel}, which provides information on the applicant and up to three accompanying persons.
Information on names, birthdates and birth places, as well as some auxiliary information on the process are given.
The general form is machine printed, while all entries are handwritten.
Formatting and writing styles are very clear and the documents show close to no degradations.
The entire collection includes $140114$ individual processes.
See \Cref{fig:mantel_distribution}, for the distribution of the number of individuals included in each process.
Note, that the form accounts only for up to four persons to be considered.
If more than four people were considered, the form was commonly continued on the back of the page.
As the dataset is designed for single full-page information extraction tasks, we removed all cover pages referring to more than four persons, leaving a total number of $135951$ pages with $203112$ individuals.

Beside the cover pages, which are comparably straight forward in terms of layout and writing styles, the benchmark includes pages from the application documents.
We extracted the first page of the application documents after the cover page.
The resulting collection shows a high variance as the application processes follow several different forms.
Focusing on the first page of the documents has the benefit that the name and the birthdate of the applicant are consistently found on the document.
Similar to the cover pages, the general forms are machine printed with handwritten entries.
The different templates, as well as the frequently occurring handwritten annotations, stamps and other additions result in a comparably high variance compared to the cover pages.

\begin{figure}[t]
  \includegraphics[width=\textwidth]{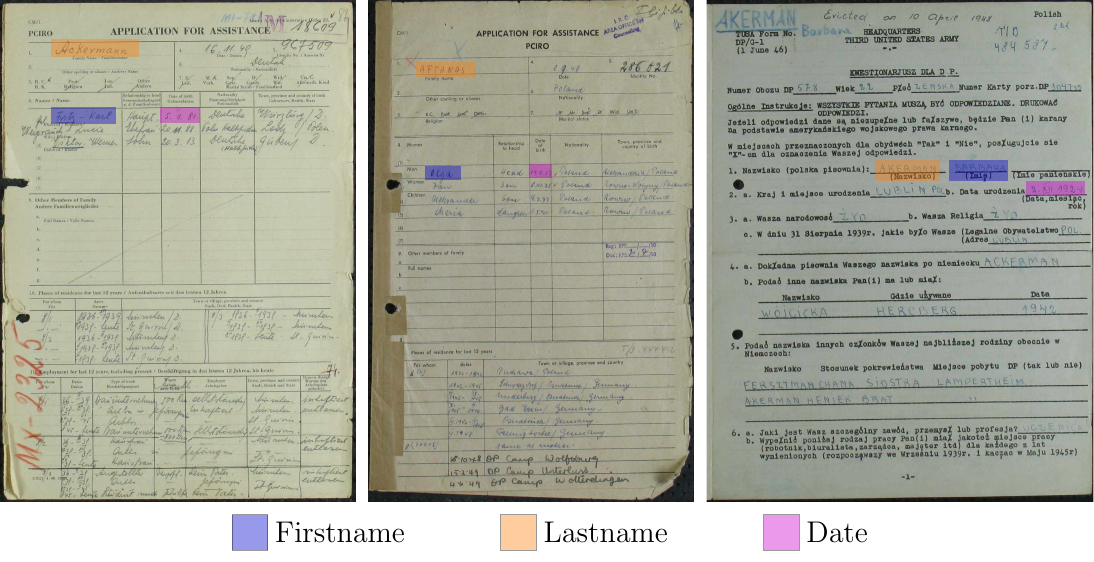}
  \caption{Example of different document pages, containing the birthdate, first and lastname of the applicant.}
  \label{fig:cover_beispiele}
\end{figure}

\begin{figure}[t]
  \includegraphics[width=\textwidth]{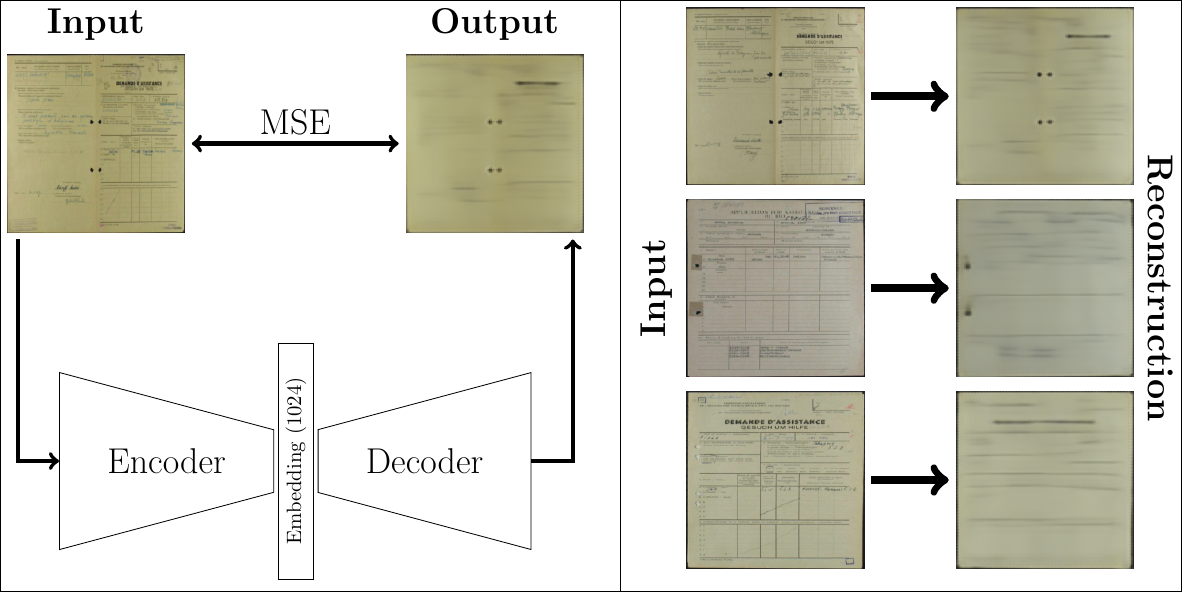}
  \caption{Left: Structure of the autoencoder that is used to generate characteristic visual embeddings. Right: Examples of image reconstruction.}
  \label{fig:autoencoder}
\end{figure}

In order to ensure a reliable foundation of documents for the benchmark, we semi-automatically cleaned the collection of documents.
Our goal was that all documents follow one of a few templates.
This scenario occurs often in many archival digitization efforts, as many bureaucratic processes of recent history relied on paper forms.
Their exact template is often unknown or hard to reliably define for a given collection.
To represent this scenario and to ensure that the target information is consistently contained in the document pages, we performed the following curation process.
A vector encoding that is representative for the visual characteristics of a document is extracted by use of an autoencoder (see \Cref{fig:autoencoder}, left).
Therefore, we first resize all document image to a resolution of 512 by 512 pixels.
The general autoencoder architecture follows the traditional encoder-decoder approach and is based on a ResNet50 \cite{He2016-RESNET}.
The size of the bottleneck, which also corresponds to the resulting size of the extracted embeddings, is $1024$.
We follow the standard approach and train the autoencoder to reconstruct the resized document images using ADAM optimization \cite{Kingma2014-ADAM} until the applied mean squared error (MSE) loss converges.
After training, the encoder is used to extract a feature representation of each image.
Looking at the reconstructed images (see \Cref{fig:autoencoder}, right), it can be observed that most of the details are lost.
This especially corresponds to the handwritten entries, while the general template persists.
It can be assumed that this characteristic translates to the extracted embeddings which encode rather template than detail information.

The resulting embeddings are, therefore, well suited to cluster the given documents with the goal to identify different document classes.
Our aim was to first generate more clusters than actual templates in the collection and later remove clusters that contain mostly outliers or documents not fitting any of the general templates.
Based on the representation provided by the autoencoeder, clustering is performed using kmeans with $50$ being the number of clusters.
After clustering, each document cluster was manually inspected.
Clusters where most documents followed a common template that includes the name and the birthday of the main applicant were kept.
Overall, nine clusters were removed from the document collection leaving $41$ clusters with a total number of $105427$ document images.
In general, the images have a comparable high resolution, while the area of interest which holds the desired information like the name or date is rather small.
This can lead to problems when models are used that allow the extraction of an information without a designated layout analysis step.
Those models usually rely on resizing the input image to fixed size.
A significant portion of the dataset consists of document images of double pages.
Clusters corresponding to double page template were identified and the corresponding images were cropped to the relevant single page.
Therefore, the resulting benchmark only includes single page images.

\begin{figure}[t]
  \includegraphics[width=\textwidth]{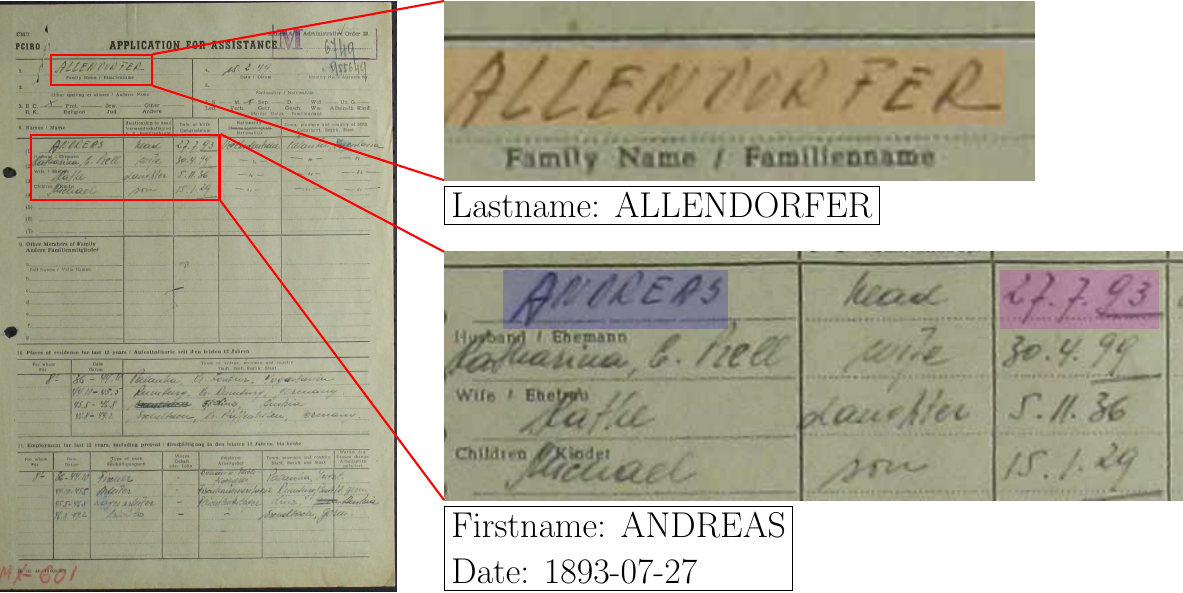}
  \caption{Example of a common name template. First- and lastname are given separately. The firstname of the main applicant occurs in a table among other family members.}
  \label{fig:cm1name}
\end{figure}

\subsection{Tasks}
Based on the cleaned document collection and the corresponding annotations, we propose two datasets and three different task for full-page information extraction.
Therefore, we introduce the three benchmarks \textit{CM1-COVER}, \textit{CM1-NAME} and \textit{CM1-DATE}.

\textit{CM1-COVER} is based on the cover pages and constitutes the task of extracting the full list of people and their corresponding birthdates.
The expected output is a string following the JSON format holding all relevant information.
In order to extract the information, the designed model has to perform multiple tasks such as segmentation and handwriting recognition making it a challenging problem to learn in an end-to-end fashion.

\textit{CM1-NAME} is based on the cropped and cleaned single-pages extracted from the application documents.
The goal is to extract the full name of the main applicant.
A main challenge in terms of document understanding is that for a big of part of the collection the first and last name do not occur consecutively.
The family name is given in a header while all corresponding persons are listed in a table later in the document.
Therefore, extracting the full name of the main person does not only mean the transcription of the name but also the identification of the main person of the document.
See \Cref{fig:cm1name} for an example.

The third benchmark, \textit{CM1-DATE}, aims at the extraction of the birthdate of the main applicant.
Again, only the first single page of the application is considered.
This is an especially important task in many digitization and publication efforts, as the birthdate of the considered people often determines if a document may be published under data protection considerations.
Most documents hold several different dates which either correspond to the documentation process or other persons on the document.
A significant challenge, therefore, is constituted by identifying which date refers to the main applicant.

For all three benchmarks, we provide training splits, a validation set comprised of $5000$ and a test set with $10000$ samples.
While in this academic context a significant number of annotated training samples can be provided, this is often not the case in an application scenario.
Evaluating model performances with respect to the availability of training data is an important consideration.
Reducing the amount of training data by training on a randomly drawn subset is a common practice, but works are often hard to reproduce and compare as the exact subsets are not published.
With few-shot learning being a main research area to consider when solving the proposed extraction tasks, we propose and publish well-defined subsets of the training data including only $1\%$, $10\%$ or $25\%$ of all samples. 
See \Cref{tab:samples} for an overview of the number of samples in the different dataset splits.

\begin{table}[t]
\caption{Number of samples per dataset split, including training subsets, validation, and test sets.}
\label{tab:samples}
\begin{tabularx}{\textwidth}{L{7} C{4.0} C{4.0} C{4.0} C{4.0} C{4.0} C{4.0} C{4.0}}
\toprule
\multirow{2}{*}{Dataset}  & \multicolumn{4}{c}{Train} & \multirow{2}{*}{Val} & \multirow{2}{*}{Test} & \multirow{2}{*}{Total} \\ [0.5ex] 
\cmidrule(ll){2-5}
                          & 1\% & 10\% & 25\% & 100\% & & & \\ [0.5ex] 
\midrule
\textit{CM1-COVER}       & $1209$ & $12095$ & $30237$ & $120951$ & $5000$ & $10000$ & $135951$ \\   
\midrule
\textit{CM1-NAME}        & \multirow{2}{*}{$904$} & \multirow{2}{*}{$9042$} & \multirow{2}{*}{$22606$} & \multirow{2}{*}{$90427$} & \multirow{2}{*}{$5000$} & \multirow{2}{*}{$10000$} & \multirow{2}{*}{$105427$}\\   
\textit{CM1-DATE}        & & &  &  & & & \\   
\bottomrule
\end{tabularx}
\end{table}


\section{Full-Page Information Extraction}
\label{sec:models}

Traditional information extraction methodologies predominantly employ sequential document analysis pipelines, consisting of independent components such as text detection, layout analysis, text recognition and semantic analysis (e.g. \cite{Tarride2023-LSG}). 
While this modular approach allows for targeted optimization of individual components, it also introduces the challenge of error propagation, where errors in early stages of processing can negatively affect downstream tasks \cite{Tarride2022-ACS,Tarride2023-LSG,Boros2020-ACO}. 
This limitation is particularly significant when dealing with complex or historical documents, which often have heterogeneous layouts, handwritten annotations and inconsistent structures \cite{Tarride2023-NER2}.

End-to-end models for full-page information extraction have emerged as a promising alternative to sequential pipelines \cite{Coquenet2023-DAS,Kim2022-OFD,Davis2022-DESSURT, Lee2023-Pix2Struct}. 
These models process entire document images within a unified framework, integrating layout analysis, text recognition and semantic analysis into a single approach. 
By taking a holistic approach, end-to-end models effectively reduce error propagation and enable consistent optimization across all processing stages \cite{Tarride2023-NER2}.
Transformer-based encoder-decoder architectures have demonstrated remarkable efficiency in this domain, offering the ability to handle diverse document formats without requiring explicit layout segmentation \cite{Kim2022-OFD}.

Large Vision Language Models (LVLMs) combining robust visual encoders with decoders based on LLMs have evolved, leading to excellent performance in many domains \cite{Yin2024_SML}. 
These models are particularly beneficial in scenarios with limited labeled training data, as they support few-shot and zero-shot learning paradigms \cite{Brown2020_LMF}. 
Recently, open-weight LVLMs were released that are comparable to commercial solutions, designed with only about 3 billion parameters, facilitating training and deployment without the need for large GPU clusters \cite{Chen2024_HFA}.

This paper evaluates both traditional and modern end-to-end models for our introduced CM1 benchmarks. 
The selected models include DONUT \cite{Kim2022-OFD} and two advanced LVLMs, PaliGemma3B \cite{Beyer2024-PAV} and Qwen2.5-VL \cite{Bai2025-QVL}, which offer distinct architectural advantages and demonstrate suitability for environments with limited computing resources.

\subsection{DONUT}
The Document Understanding Transformer (DONUT) is a Transformer-based encoder-decoder model specifically optimized for full-page information extraction. 
The architecture of DONUT consists of a Swin Transformer \cite{Liu2021-SWIN} as the encoder and a BART (Bidirectional and Auto-Regressive Transformers) \cite{Lewis2020-BART} model as the decoder. 
The Swin Transformer employs a hierarchical feature extraction approach using shifted windows, effectively capturing both local and global visual features from document images. 
This capability is critical for processing high-resolution input and managing complex document layouts. 
The BART decoder uses an autoregressive text generation approach, enabling DONUT to produce structured output, such as JSON-based key-value pairs, directly from raw document images.

The pre-training strategy for DONUT involves large-scale text recognition tasks using both real-world document images with pseudo-labeled OCR output and synthetically generated document images. 
This dual pre-training approach enhances the generalization capabilities of the model across different text styles and formats, including both printed and handwritten content. 
With approximately 201 million parameters, DONUT represents a balanced compromise between model performance and resource efficiency, allowing fine-tuning on standard hardware without the need for large-scale computing infrastructure.

\subsection{PaliGemma}
PaliGemma3B is an open-weight LVLM that integrates the SigLIP-So400m vision encoder \cite{Zhai2023-SIG} with the Gemma-2B language model \cite{Mesnard2024-GEMMA}, resulting in a model with approximately 3 billion parameters. 
The architecture uses a Vision Transformer (ViT) \cite{Dosovitskiy2021-AII}  encoder to process input images and extract high-dimensional visual features, which are then linearly projected to match the input space of the Gemma-2B language model. 
This integrated architecture facilitates the seamless processing of multimodal input, allowing the model to effectively handle tasks requiring both visual understanding and natural language generation.

The pre-training of PaliGemma follows a multi-stage approach to cultivate robust unimodal and multimodal competencies. 
First, the SigLIP-So400m vision encoder and the Gemma-2B language model are independently pre-trained on large unimodal datasets to improve the feature extraction and natural language processing capabilities of the model, respectively. 
The model then undergoes multimodal pre-training on a variety of image-text pairs, focusing on tasks such as image captioning and visual question answering. 
This phase enables the model to learn complex relationships between visual elements and textual descriptions, which is critical for performance in integrated vision-language tasks.

PaliGemma supports a range of input image resolutions, with the highest resolution of 896px used in this work to capture detailed visual information. The model used has been pre-trained on the DocVQA dataset \cite{Mathew2021-DOCVQA}, providing it with advanced capabilities for document-based visual question answering.

\subsection{Qwen2.5-VL}
Qwen2.5-VL is a state-of-the-art LVLM based on a fully Transformer-based architecture that combines a ViT \cite{Dosovitskiy2021-AII} with the Qwen2.5 LLM \cite{Yang2024-QLL}. The encoder uses a dynamic resolution ViT with window attention \cite{Dehghani2023-PnP}, which allows the model to efficiently handle varying input image sizes and produces a variable number of vision tokens.
This dynamic image resolution allows Qwen2.5-VL to process images at their native resolution, which is particularly beneficial when interpreting complex documents and extracting detailed visual information.
The embedding of the visual tokens is adapted to the input size of the LMM by cross-modal attention \cite{Bai2025-QVL}.

The Qwen2.5 LLM is pre-trained on 18 trillion tokens, leading to high performance on tasks requiring deep semantic understanding and high quality text generation \cite{Yang2024-QLL}.
The model is designed for a wide range of multimodal tasks, including object localisation, robust document understanding and video analysis \cite{Bai2025-QVL}. 
The model has been trained on diverse multimodal datasets, including not only natural images and synthetic visual data, but also various structured document images such as invoices, forms, tables and complex layouts. 
This comprehensive pre-training approach enables Qwen2.5-VL to state-of-the-art performances on several document image analysis benchmarks \cite{Bai2025-QVL}.

\subsection{Training (PEFT)}
To adapt both LVLMs to the specific requirements of historical document understanding, we employ Parameter Efficient Fine-Tuning (PEFT) techniques, particularly Low-Rank Adaptation (LoRA) \cite{Hu2022_LLR} and Quantized LoRA (QLoRA) \cite{Dettmers2023_QEF}. 
LoRA has demonstrated a strong approximation to full fine-tuning of LLMs, achieving comparable performance across various tasks while significantly reducing computational and memory demands \cite{Hu2022_LLR}. 
By introducing low-rank matrices into the training of LVLMs, LoRA allows for efficient fine-tuning by modifying only a subset of parameters. 
QLoRA further enhances this efficiency by applying quantization techniques that reduce the memory footprint through lower-precision weight representations. 
These approaches substantially decrease computational costs, enabling fine-tuning on single-GPU setups and broadening access to high-performance LVLMs in resource-constrained environments \cite{Hu2022_LLR,Dettmers2023_QEF}.
Our fine-tuning strategy targets both the encoder and decoder components of the models, improving their ability to accurately interpret the high variability of document images. 

\section{Experiments}
\label{sec:experiments}
To evaluate the performance of the proposed approaches from the previous section for full-page information extraction, we conducted a series of experiments using the CM1 dataset. 
Thereby, DONUT, PaliGemma, and Qwen2.5-VL were trained and tested using the predefined training, validation, and test sets. 
Additionally, to establish a baseline for general-purpose models, we performed a zero-shot evaluation using ChatGPT-4o. 
Since ChatGPT-4o  has presumably not been fine-tuned on CM1 datasets, this evaluation serves as a reference point to understand how a highly capable vision-language model performs without any adaptation.

\subsection{Training Setup}
The training was conducted on a single NVIDIA A100 GPU with 80GB VRAM, ensuring sufficient computational resources for fine-tuning large-scale models. 
We employed the AdamW optimizer \cite{Loshchilov2019-ADAMW} with an initial learning rate of $1e^{-4}$ for PaliGemma\footnote[2]{PaliGemma: google/paligemma-3b-ft-docvqa-896} and Qwen2.5-VL\footnote[3]{Qwen2.5-VL: Qwen/Qwen2.5-VL-3B-Instruct} while we use a rate of $3e^{-5}$ for DONUT\footnote[4]{DONUT: naver-clova-ix/donut-base}, following a cosine annealing learning rate schedule to progressively reduce the learning rate as training progressed.
For our experiments, we utilized LoRA for LVLMs with a rank of 8 and an alpha of 16 for all linear layers of the models, a configuration that balances efficiency and adaptability. 
To further reduce memory consumption, QLoRA was applied to Qwen2.5-VL, enabling quantized weight updates, thereby making fine-tuning feasible on a single GPU.
We adopted different batch size configurations, using a batch size of 1 with batch aggregation of 4 for LVLMs and a batch size of 4 with batch aggregation of 4 for DONUT. The image input for QwenVL was resized to a minimum of 256 and a maximum of 1280 pixels to accommodate high-resolution inputs during training. 
For text generation tasks, the maximum token length was set to 32 for \textit{CM1-DATE} and \textit{CM1-NAME} tasks and 512 for \textit{CM1-COVER}.
Additionally, we applied a warmup phase of 10,000 iterations for DONUT to stabilize training. 
The maximum image input dimensions of DONUT were set to 2560 $\times$ 1920 pixels, with a sequence length cap of 768.

\subsection{Metrics}
To assess the performance of the models on our proposed benchmarks, we use the Character Error Rate (CER), accuracy and Tree Edit Distance (TED).
Thereby, the CER measures the edit distance at the character level between the extracted and ground-truth text.
A lower CER value indicates better text recognition performance, with 0\% CER meaning a perfect extraction. 
This metric is particularly useful in evaluating OCR-based errors, where minor character-level mismatches can accumulate, affecting the extracted text quality.

The accuracy evaluates the percentage of correctly extracted entities (e.g., full names, birthdates) that match the ground truth exactly. 
This metric is particularly important for real-world applications where partial extraction correctness is insufficient.

The TED is inspired by \cite{Kim2022-OFD} and used for assessing the quality on \textit{CM1-COVER}.
This metric measures how well the extracted structured information matches the expected format. 
To compute TED, the extracted output is first converted into a tree representation, ensuring that the hierarchical structure aligns with the ground truth. 
The edit distance between the two trees is then computed, accounting for insertions, deletions, and modifications of nodes. 
To make the score comparable across different documents, the computed tree distance is normalized by the size of the ground-truth tree. 
The final TED accuracy score is derived using:
\begin{equation}
	TED_{\text{acc}} = \max\left(0, 1 - \frac{d(T_1,T_2)}{|T_2|}\right)
\end{equation}
where $d(T_1, T_2)$ represents the distance between the predicted tree $T_1$ and the ground-truth tree $T_2$, and $|T_2|$ is the size of the ground-truth tree.

Each of these metrics provides a different perspective on model performance. 
Thereby, CER measures OCR quality, accuracy determines whether entire entities are correctly extracted and TED evaluates structured data alignment. 

\subsection{Results}
The evaluation of full-page information extraction was conducted using the three proposed benchmarks: cover page extraction (\textit{CM1-COVER}), name extraction (\textit{CM1-NAME}), and birthdate extraction (\textit{CM1-DATE}). 
Each task has the challenges of structured field identification and handling handwritten text. 
In the following, we analyze the results of our experiments and compare the performance of DONUT, PaliGemma3B, and Qwen2.5-VL, as well as discuss the zero-shot baseline performance of ChatGPT-4o.

\subsubsection{Cover Page Extraction}

The \textit{CM1-COVER} benchmark involves extracting structured key-value pairs, specifically names and birthdates, from a predefined document layout. 
Given this structured format, models can effectively leverage layout information to enhance extraction accuracy.
As given in \Cref{tab:cm1_cover}, all models successfully learned to generate the required structured output within a few iterations, demonstrating their ability to adapt quickly to the expected format. However, performance varied depending on the amount of available training data.
DONUT initially struggled in low-data settings, exhibiting high CERs and low TED accuracy when trained on only 1\% of the dataset. 
However, with increased training data, its performance improved rapidly, surpassing both PaliGemma3B and Qwen2.5-VL at 10\%, 25\%, and 100\% training data, ultimately achieving the best overall results.
In contrast, the LVLMs showed stronger few-shot capabilities with 1\% of the training data. 
However, as training data increased, the performance gap between the models narrowed, with results becoming increasingly comparable across all settings.

\begin{table}[t]
  \caption{Comparison on the \textit{CM1-COVER} benchmark for different training data sizes.}
  \label{tab:cm1_cover}
\begin{tabularx}{\textwidth}{L{1.3} C{0.2} C{0.4}C{0.4} C{0.2} C{0.4}C{0.4} C{0.2} C{0.4}C{0.4} C{0.2} C{0.4}C{0.4} }
\toprule
\multirow{2}{*}{Model} & & \multicolumn{2}{c}{1\%} & & \multicolumn{2}{c}{10\%}& & \multicolumn{2}{c}{25\%} & & \multicolumn{2}{c}{100\%} \\ [0.5ex] 
\cmidrule(ll){3-4}\cmidrule(ll){6-7} \cmidrule(ll){9-10} \cmidrule(ll){12-13}
                       & & TED & CER & & TED & CER & & TED & CER & & TED & CER\\
	\midrule
  Donut    &&  25.6   & 53.5 &&  \textbf{88.2}  &  \textbf{8.9}  &&   \textbf{90.4} &  \textbf{6.9}   &&    \textbf{92.4}   & \textbf{6.9}   \\ 
  PaliGemma        & & \textbf{75.5} & \textbf{13.3} & & $83.2$ & $11.3$ & & $88.0$ & $8.3$ & & $90.1$  & $8.4$ \\   
  Qwen2.5-VL         & & 58.8 & 15.9 & &  87.0   &  9.1   &&   89.3  &  8.1   &&  89.8   &  7.8  \\  
\bottomrule
\end{tabularx}

\vspace*{0.2em}
{\scriptsize TED ($\uparrow$) and CER ($downarrow$) scores are multiplied by 100}
\end{table}

\subsubsection{Name Extraction}
The \textit{CM1-NAME} benchmark presents a more complex challenge, as the first and last names of individuals are often not located sequentially within the document. 
This task requires the models to learn both the visual structure and logical relationships between different text fields.
\Cref{tab:cm1_name} shows that DONUT struggles significantly in low-data settings, failing to extract names at 1\% and 10\% training data. 
However, with more training data, its accuracy improved rapidly, outperforming Qwen2.5-VL and PaliGemma3B at 100\% training data.
In contrast, Qwen2.5-VL demonstrated superior few-shot performance, achieving the highest accuracy at 1\% and 10\% training data, while PaliGemma3B also performed well in low-data scenarios. 
As training data increased, performance differences between the models narrowed.

\begin{table}[t]
\caption{Comparison on the \textit{CM1-NAME} benchmark for different training data sizes.}
\label{tab:cm1_name}
\begin{tabularx}{\textwidth}{L{1.3} C{0.2} C{0.4}C{0.4} C{0.2} C{0.4}C{0.4} C{0.2} C{0.4}C{0.4} C{0.2} C{0.4}C{0.4}}
\toprule
\multirow{2}{*}{Model} & & \multicolumn{2}{c}{1\%} & & \multicolumn{2}{c}{10\%}& & \multicolumn{2}{c}{25\%} & & \multicolumn{2}{c}{100\%} \\ [0.5ex]
\cmidrule(ll){3-4}\cmidrule(ll){6-7} \cmidrule(ll){9-10} \cmidrule(ll){12-13} 
                       & & Acc & CER & & Acc & CER & & Acc & CER & & Acc & CER\\
	\midrule

	Donut            &&  0.0  & 93.5 &&  0.0   &  85.1   &&   53.5  & 15.1    &&  \textbf{61.1}    &  \textbf{12.3}  \\   
  PaliGemma        & & $26.5$ & $27.3$ & & $40.3$ & $18.7$ & & $47.0$ & $15.7$ & & $54.3$ & $13.3$ \\   
	Qwen2.5-VL         &&  \textbf{35.1}   & \textbf{20.8} &&   \textbf{46.0}  &  \textbf{17.0}   &&  \textbf{53.8}   & \textbf{13.8}    &&  55.2  &  12.5  \\
\bottomrule
\end{tabularx}

\vspace*{0.2em}
{\scriptsize Acc ($\uparrow$) and CER ($\downarrow$) scores are multiplied by 100}
\end{table}

\subsubsection{Birthdate Extraction}

The \textit{CM1-DATE} benchmark posed a significant challenge, as historical documents often contain multiple dates, making it difficult for models to correctly identify the birthdate of the main applicant, as given in \Cref{tab:cm1_date}.
As in previous benchmarks, DONUT struggled in low-data settings, achieving near-zero accuracy at 1\%, 10\%, and 25\% training data, likely due to overfitting, before improving sharply at 100\% training data.
Qwen2.5-VL significantly outperformed PaliGemma3B at 1\% training data, demonstrating superior few-shot generalization. While this advantage persisted across training levels, the gap narrowed, with DONUT being the best model with full training data.

\begin{table}[t]
\caption{Comparison on the \textit{CM1-DATE} benchmark for different training data sizes.}
\label{tab:cm1_date}
\begin{tabularx}{\textwidth}{L{1.3} C{0.2} C{0.4}C{0.4} C{0.2} C{0.4}C{0.4} C{0.2} C{0.4}C{0.4} C{0.2} C{0.4}C{0.4} }
\toprule
\multirow{2}{*}{Model} & & \multicolumn{2}{c}{1\%} & & \multicolumn{2}{c}{10\%}& & \multicolumn{2}{c}{25\%} & & \multicolumn{2}{c}{100\%} \\ [0.5ex] 
\cmidrule(ll){3-4}\cmidrule(ll){6-7} \cmidrule(ll){9-10} \cmidrule(ll){12-13}
                       & & Acc & CER & & Acc & CER & & Acc & CER & & Acc & CER\\
	\midrule
	Donut            && 0.0 & 47.9 && 0.1 & 48.4 && 0.1 & 48.7 && \textbf{77.7} & \textbf{7.6} \\ 
  PaliGemma        & & $24.3$ & $22.1$ & & $53.0$ & $12.2$  & & $59.2$ & $10.5$ & & $66.5$ & $9.0$ \\   
	Qwen2.5-VL        &&   \textbf{48.1}  & \textbf{14.0} &&  \textbf{59.7}   &  \textbf{11.0}   &&  \textbf{64.6}    &  \textbf{9.8}   &&  69.3   &  8.4  \\   
\bottomrule
\end{tabularx}\\

\vspace*{-0.2em}
{\scriptsize Acc ($\uparrow$) and CER ($\downarrow$) scores are multiplied by 100}
\end{table}

\subsubsection{Zero Shot Evaluation}
This experiment aims to evaluate whether a state-of-the-art vision-language model could extract key information from historical documents without prior training on CM1. ChatGPT-4o\footnote[5]{Model-ID: gpt-4o-2024-08-06} was tested in a zero-shot setting using the official OpenAI API with the following prompts:
\begin{itemize}
  \item {\small\textbf{CM1-COVER}:} Extract the first name, surname and date of birth of each person from this image and return this information in JSON format. Example:  \{'Person': [\{'firstname': 'Doe', 'lastname': 'John', 'Geb-Dat': '1913-01-31'\}, \{'firstname': 'Doe', 'lastname': 'Jane', 'Geb-Dat': '1923-04-31'\}]\}
  \item {\small\textbf{CM1-NAME}:} Extract only the first and last name of the main person (head) from this image and output only the name without any additional text.
  \item {\small\textbf{CM1-DATE}:} Extract only the date of birth of the main person (head) from this image and output only the date of birth in the format yyyy-mm-dd without any additional text.
\end{itemize}

The zero-shot evaluation in \Cref{tab:zero_shot} highlights key differences in model capabilities. 
DONUT shows limited success in handling structured document tasks but fails at entity extraction, likely due to its pre-training focus. 
PaliGemma3B, not optimized for these tasks, struggles to extract meaningful information.
Qwen2.5-VL, despite being a 3B parameter open-weight model, demonstrates strong document understanding, producing well-structured outputs and handling entity extraction. 
ChatGPT-4o, a significantly larger commercial model, performs best overall but often returns fallback text instead of precise extractions. 
The competitive performance of Qwen2.5-VL underscores the potential of smaller open-weight models as viable alternatives to large-scale proprietary solutions.

\begin{table}[t]
	\caption{Zero-shot performance on the CM1 benchmarks.}
	\label{tab:zero_shot}
	\begin{tabularx}{\textwidth}{L{1} C{0.2} C{0.4}C{0.4} C{0.2} C{0.4}C{0.4} C{0.2} C{0.4}C{0.4} }
		\toprule
		\multirow{2}{*}{Model} & & \multicolumn{2}{c}{\textit{CM1-COVER}} & & \multicolumn{2}{c}{\textit{CM1-NAME}}& & \multicolumn{2}{c}{\textit{CM1-DATE}} \\ [0.5ex] 
		\cmidrule(ll){3-4}\cmidrule(ll){6-7} \cmidrule(ll){9-10}
		& & TED & CER & & Acc & CER & & Acc & CER \\
		\midrule
		Donut  &&  13.0  &  89.0  &&  0.0   &  100.0  &&   0.0   &   100.0 \\
		PaliGemma  &&  0.0  &  98.2  &&   0.0  &  96.4  &&   0.0   &  77.9  \\
		Qwen2.5-VL  &&  41.1  & 53.6  &&  8.2   &  \textbf{63.1}  &&   13.7   &  42.7  \\
		ChatGPT4o  && \textbf{58.8}  &  \textbf{22.6}  &&   \textbf{22.9}  &  79.9   &&   \textbf{26.3}   & \textbf{28.1}   \\
		\bottomrule
	\end{tabularx}
	
	\vspace*{0.2em}
	{\scriptsize  TED ($\uparrow$), Acc ($\uparrow$) and CER ($\downarrow$) scores are multiplied by 100}
\end{table}

\subsubsection{Discussion}
The results across all three benchmarks highlight key insights regarding pre-training, fine-tuning, and structured information extraction. 
Models with large-scale pre-training, such as PaliGemma and Qwen2.5-VL, demonstrate superior generalization capabilities in few-shot settings, consistently outperforming DONUT when trained on limited data. 
This suggests that multimodal pre-training plays a crucial role in learning robust feature representations. 
However, full fine-tuning remains essential for optimal performance, as DONUT achieves competitive results only when trained on the full dataset. 

The zero-shot evaluation of GPT-4o demonstrates promising structural extraction capabilities, correctly predicting about a quarter of names and birthdates without prior exposure to CM1. 
However, fallback responses and handwriting misinterpretations significantly impact CER, reinforcing the need for fine-tuning to improve field association and handwriting robustness.

\section{Conclusions}
In this work, we present a novel dataset and benchmarks to evaluate end-to-end information extraction from full-page documents.
In our experiments, we compare a traditional full-page extraction model to two open-weight LVLMs.
The results show that the traditional model, despite being of significantly smaller size, outperforms the LVLMs when enough training data is available.
Nonetheless, this scenario is unlikely in a practical application.
When only few annotated samples can be used for training, the LVLMs strongly benefit from their intensive pretraining and offer a viable alternative. 
Furthermore, our results show that despite being successful few-shot learners, zero-shot performances are still inadequate for a historic collection containing handwriting.
This is also the case for a commercial model such as ChatGPT4o.

\newpage

%
%
%
\bibliographystyle{splncs04}
\bibliography{literature}

\end{document}